# A computer vision-based model for occupancy detection using low-resolution thermal images

Xue Cui [1], Vincent Gbouna Zakka [2], Minhyun Lee [3],

[1]Department of Building and Real Estate, The Hong Kong Polytechnic University, Hong Kong, China, xue.cui@connect.polyu.hk
[2] College of Engineering and Physical Sciences, Aston University, United Kingdom, vzakk22@aston.ac.uk
[3] Department of Building and Real Estate, Research Institute for Smart Energy (RISE), The Hong Kong Polytechnic University, Hong Kong, China, minhyun.lee@polyu.edu.hk

**SUMMARY**
Occupancy plays an essential role in influencing the energy consumption and operation of heating, ventilation, and air conditioning (HVAC) systems. Traditional HVAC typically operate on fixed schedules without considering occupancy. Advanced occupant-centric control (OCC) adopted occupancy status in regulating HVAC operations. RGB images combined with computer vision (CV) techniques are widely used for occupancy detection, however, the detailed facial and body features they capture raise significant privacy concerns. Low-resolution thermal images offer a non-invasive solution that mitigates privacy issues. The study developed an occupancy detection model utilizing low-resolution thermal images and CV techniques, where transfer learning was applied to fine-tune the You Only Look Once version 5 (YOLOv5) model. The developed model ultimately achieved satisfactory performance, with $precision$, $recall$, $mAP50$, and $mAP50-95$ values approaching 1.000. The contributions of this model lie not only in mitigating privacy concerns but also in reducing computing resource demands.

## 1. INTRODUCTION

The building sector is responsible for 34% of global energy consumption (United Nations Environment Programme, 2024). Heating, ventilation, and air conditioning (HVAC) systems are the most energy-intensive systems in buildings, accounting for 38% of building energy consumption (González-Torres et al., 2022). Existing HVAC systems usually operate on fixed schedules, easily causing over-conditioning (Wu et al., 2024). Occupant-centric control (OCC) has emerged as an advanced strategy to address this issue by utilizing actual occupant information to control HVAC operations. Among various occupant-related factors, occupancy plays an essential role in determining the on/off status of HVAC systems. Studies have shown that integrating occupancy into HVAC operations can achieve an energy-saving ratio between 19% and 44% (Pang et al., 2020), demonstrating the significant benefits of leveraging occupancy for HVAC control.

Incorporating occupancy into HVAC operations begins with occupancy detection, which forms the essential foundation. Existing studies primarily rely on passive infrared, radio frequency signal technology, cameras, environmental sensors, Wi-Fi, and Bluetooth for occupancy detection (Ding et al., 2022). Passive infrared, radio frequency signal technology, and environmental sensors are relatively easy to deploy (Ding et al., 2022). However, passive infrared is unable to detect static people and multi-occupant scenarios, and has a limited detection range (Choi et al., 2021; Ding et al., 2022). Radio frequency signal technology is highly sensitive to indoor electromagnetic conditions (Ding et al., 2022), while environmental sensors (e.g., $CO_2$ sensors) often experience delay issues when detecting occupancy (Choi et al., 2021). Wi-Fi and Bluetooth are efficient and convenient, however, their connection points do not always match the actual number of occupants (Ding et al., 2022). Cameras offer high accuracy in detecting occupancy and can immediately track changes in occupants (Choi et al., 2021; Ding et al., 2022), in particular, computer vision (CV) techniques are commonly applied to images for occupancy detection (Choi et al., 2021).

Many existing CV-based occupancy detection models employ Red, Green, and Blue (RGB) images for detecting occupancy. Tien et al. (2020b) applied a convolutional neural network to images to detect occupancy activity, achieving an average detection accuracy of 80.62%. Similarly, Choi et al. (2021) integrated a CV-based occupancy-centric counting method into the HVAC and lighting systems in two offices, achieving a 10.2% energy reduction. Overall, CV-based occupancy detection models commonly achieve high accuracy while contributing to building energy savings. However, most of these studies rely on high-resolution RGB images, which raise significant privacy concerns since the detailed facial and body features of occupants are captured (Choi et al., 2021). In addition, the high-resolution nature of RGB images requires high computing power and memory capacity (O'Mahony et al., 2020). To address this challenge, it is crucial to conserve computing resources when developing a CV-based occupancy detection model. Here, low-resolution thermal images provide an effective option since they not only mitigate privacy concerns but also reduce processing demands.

In summary, existing occupancy detection models reveal two main research gaps. First, most occupancy detection approaches based on passive infrared, radio frequency signal technology, environmental sensors, Wi-Fi, and Bluetooth, encounter significant limitations in practical applications. Second, the use of RGB images for occupancy detection raises privacy concerns and requires high computing resources. To address these gaps, this study aims to develop an occupancy detection model leveraging low-resolution thermal images and CV techniques. The novelty lies in its non-invasive nature and ability to mitigate privacy concerns, making it more psychologically acceptable to occupants. Furthermore, it reduces the requirement for computing resources, enhancing its practicality for real-world building environments.

## 2. METHODOLOGY

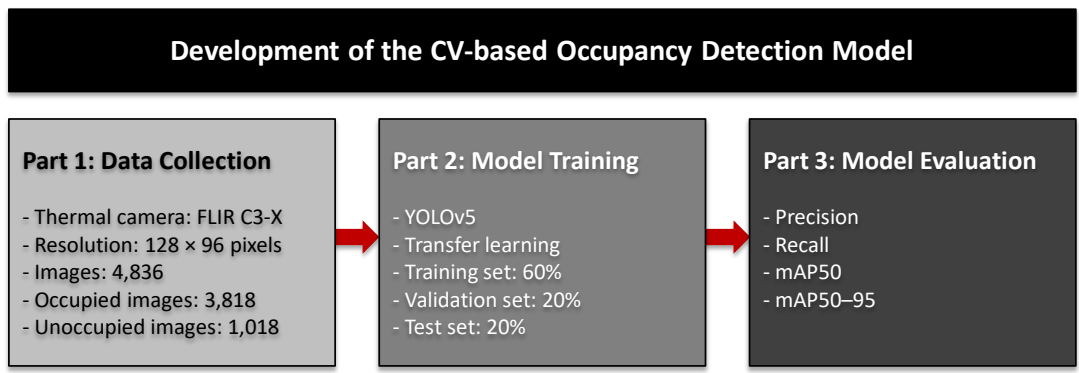

**Figure 1.** The methodological framework for developing the CV-based occupancy detection model.

Figure 1 illustrates the methodological framework for developing the CV-based occupancy detection model, consisting of three main parts: (1) data collection, which involves gathering sufficient low-resolution thermal images under occupied and unoccupied scenarios; (2) model training, where transfer learning is applied to fine-tune the You Only Look Once version 5 (YOLOv5) model using the collected thermal images; and (3) model evaluation, which assesses the reliability of the developed model based on four metrics.

## 2.1. Data collection

Occupant thermal images were collected in a single-person office at The Hong Kong Polytechnic University on 8 August 2024. To simulate a realistic working environment, a single occupant performed typical computer-based tasks following their natural working habits, without any prescribed activity schedule or arrangement. Thermal images were captured using a FLIR C3-X compact thermal camera, featuring a low-resolution of 128 × 96 pixels ("FLIR C3-X Compact Thermal Camera with Cloud Connectivity and Wi-Fi | Teledyne FLIR," 2024). The camera was mounted on the upper bezel of the computer monitor and connected to the computer via USB. To ensure sufficient data collection, thermal images were automatically captured at 10-second intervals using radiometric streaming through the FLIR Desktop Atlas SDK. By the end of the day, a total of 4,836 thermal images were collected, comprising 3,818 images with the occupant present and 1,018 without, resulting in an occupied-to-unoccupied ratio of 3.75.

## 2.2 Model training

The study initially employed the YOLOv5 model to detect occupant faces since it demonstrates efficient and robust performance in this field (Jeoung et al., 2023). However, the publicly available YOLOv5 model produced unsatisfactory detection results when applied to the low-resolution thermal images collected in this study. This result may be attributed to the YOLOv5 model being pretrained on the Common Objects in Context (COCO) dataset (Choi et al., 2021), which offers limited exposure to low-resolution thermal images. Consequently, the pretrained YOLOv5 model likely struggled to capture the distinctive features of occupant faces in low-resolution thermal images, resulting in its suboptimal detection performance.

To improve the detection accuracy of the YOLOv5 model on low-resolution thermal images, an end-to-end transfer learning approach was employed. In this approach, the original YOLOv5 weights were fine-tuned using a curated dataset of low-resolution thermal images. Unlike traditional transfer learning methods that freeze early layers to retain generic feature representations, this approach involved fine-tuning the entire model. The layers of YOLOv5 were retrained on the curated dataset of low-resolution thermal images. This end-to-end retraining ensured that the model adapted comprehensively to the specific characteristics of thermal images, such as reduced resolution, lack of color information, and distinct thermal patterns.

LabelImg, a graphical image annotation tool, was utilized to label face bounding boxes in the thermal images ("HumanSignal/labelImg," 2024). The annotations were saved in YOLO-compatible .txt files, capturing the coordinates of the labeled bounding boxes ("HumanSignal/labelImg," 2024). All thermal images containing occupants were labeled, while annotation files for unoccupied images were left blank. After completing the annotation process, the thermal images and their corresponding annotation files were randomly split into a training subset (60%), a validation subset (20%), and a test subset (20%) for model training. Table 1 presents the sample sizes for each subset, with the occupied-to-unoccupied ratio in each subset matching the overall dataset ratio of 3.75.

Table 1. Sample sizes for the training, validation, and test subsets.

| Subset | Thermal images | | Annotation labels | | Occupied-to-unoccupied ratio |
|---|---|---|---|---|---|
| | Occupied | Unoccupied | Occupied | Unoccupied | |
| Training | 2290 | 610 | 2290 | 610 | |
| Validation | 764 | 204 | 764 | 204 | 3.75 |
| Test | 764 | 204 | 764 | 204 | |

## 2.3 Model evaluation

To evaluate the reliability of the trained model, the test subset was employed to assess detection performance under unseen scenarios. In this evaluation, it is essential to set an appropriate confidence threshold, as detection accuracy is highly sensitive to this parameter (Choi et al., 2022). Specifically, only objects with confidence scores exceeding the threshold are detected, while those below it are ignored (Choi et al., 2022). Thus, a low confidence threshold raises the likelihood of falsely detecting non-occupant objects, while a high confidence threshold increases the risk of missing true occupants (Choi et al., 2022). Following prior studies (Sohn et al., 2020; Xiao et al., 2021), this study adopted a confidence threshold of 0.9 to ensure robust detection performance. In addition, four criteria were used to assess the model performance, including $precision$, $recall$, $mAP50$, and $mAP50-95$. $Precision$ and $recall$, derived from $true\ positive\ (TP)$, $false\ positive\ (FP)$, and $false\ negative\ (FN)$, are defined by equations (1) and (2) ("formula of Box(P R mAP50 mAP50-95) Issue #9221," 2024). $mAP50$ and $mAP50-95$ are based on $average\ precision\ (AP)$ calculated at specific Intersection over Union (IoU) thresholds ("formula of Box(P R mAP50 mAP50-95) Issue #9221," 2024). $mAP50$ calculates the $AP$ at an IoU threshold of 50%, while $mAP50-95$ provides a stricter assessment by averaging the $AP$ across IoU thresholds ranging from 50% to 95% in 5% increments ("formula of Box(P R mAP50 mAP50-95) Issue #9221," 2024). Higher values across these four criteria indicate better model performance ("What is Box(P R mAP50 mAP50-95)?," 2024).

$$Precision = \frac{TP}{TP + FP} \qquad (1)$$

$$Recall = \frac{TP}{TP + FN} \qquad (2)$$

## 3. RESULTS AND DISCUSSION
### 3.1. Description of thermal images

As indicated in section 2.1, a total of 4,836 thermal images were collected. Figure 2 presents representative thermal images under occupied and unoccupied scenarios. Specifically, Figure 2(a) depicts the most typical unoccupied condition, while Figure 2(b)-(j) illustrate various occupied scenarios. Among these, Figure 2(b) shows the most typical occupied condition, featuring a frontal face view. Figure 2(c)-(d) display partial face visibility, Figure 2(e)-(g) show the face obscured by a hand or bottle, and Figure 2(h)-(j) present side-face views with varying head orientations. Notably, the diversity of head orientations, occupant gestures, and the limited visibility of occupant faces in these thermal images poses significant challenges for developing an effective occupancy detection model, as further discussed in section 3.3.

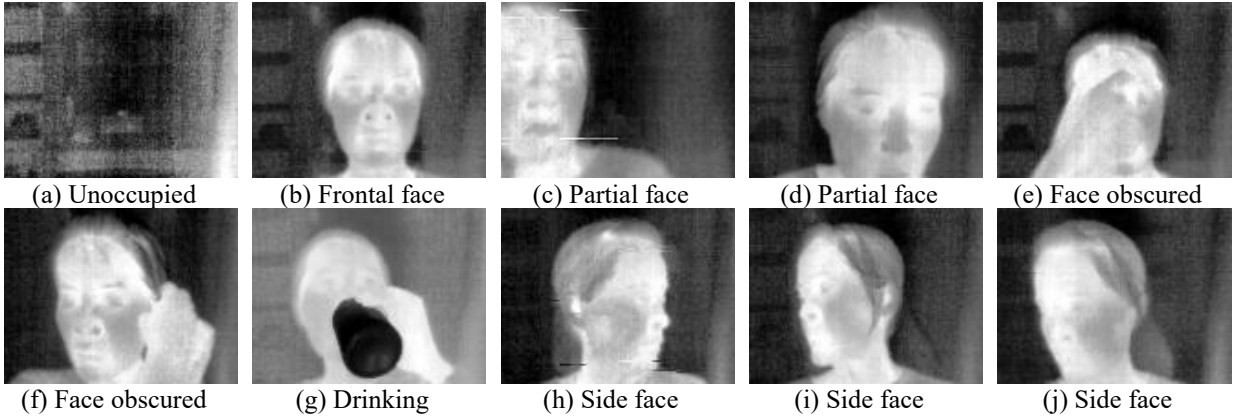

(a) Unoccupied (b) Frontal face (c) Partial face (d) Partial face (e) Face obscured
(f) Face obscured (g) Drinking (h) Side face (i) Side face (j) Side face

**Figure 2.** Representative thermal images featuring occupied and unoccupied scenarios.

### 3.2. Evaluation of model performance

Figure 3 illustrates the training curves of $precision$, $recall$, $mAP50$, and $mAP50-95$ of the model on the validation subset across 250 epochs. $Precision$ quickly converged to nearly 1.000, while $recall$ gradually stabilized around 1.000 despite minor fluctuations. Similarly, $mAP50$ rapidly stabilized near 0.995, and $mAP50-95$ progressively converged to approximately 0.980, demonstrating the robust detection capabilities of the model across varying IoU thresholds. As described in section 2.3, the study evaluated the model performance on the test subset. Table 2 summarizes the results of the four evaluation criteria on the test subset. Overall, the model performed excellently on the test subset, achieving a $precision$ of 1.000, $recall$ of 0.984, $mAP50$ of 0.991, and $mAP50-95$ of 0.975. These consistently high scores indicate the strong accuracy and generalizability of the developed occupancy detection model under unseen scenarios.

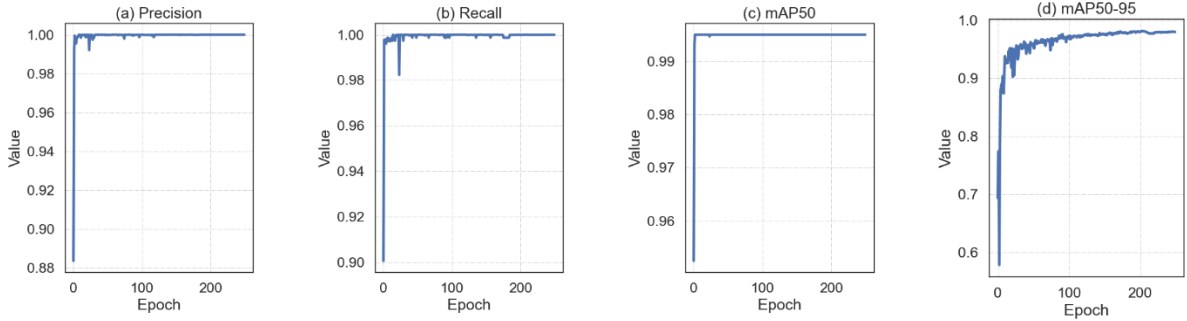

**Figure 3.** Training curves of $precision$, $recall$, $mAP50$, and $mAP50-95$ of the model on the validation subset across 250 epochs.

**Table 2.** Results of four evaluation criteria on the test subset.

|  | $Precision$ | $Recall$ | $mAP50$ | $mAP50-95$ |
|---|---|---|---|---|
| On the test subset | 1.000 | 0.984 | 0.991 | 0.975 |

To ensure the reliability of the developed occupancy detection model, the study compared its performance with other similar CV-based detection models reported in previous studies. $Precision$ and $recall$ were selected as the key comparison criteria since they are widely used to evaluate model performance in this field. Table 3 summarizes the comparison results of $precision$ and $recall$ of the developed model and other similar CV-based detection models from previous studies. The prior models examined in Table 3 primarily focus on detecting occupant activities (such as sitting, standing, and walking), the on/off status of the personal computer monitor, the open/closed status of windows, and worker tracking in off-site construction. The $precision$ of these CV-based models ranges from 0.864 to 0.987, while their $recall$ ranges from 0.828 to 0.967. By comparison, the occupancy detection model developed in this study achieved a $precision$ of 1.000 and a $recall$ of 0.984, exceeding the highest values reported among the examined models. These findings highlight the superior performance and reliability of the developed model in detecting occupancy.

Table 3. The comparison results of $precision$ and $recall$ of the occupancy detection model developed in this study and other similar CV-based detection models in previous studies.

|  | Detection target | Precision | Recall |
|---|---|---|---|
| **This study** | **Occupancy** | **1.000** | **0.984** |
| (Tien et al., 2020a) | Occupant activity | 0.932 | 0.930 |
|  | Personal computer monitor _ on | 0.914 | 0.860 |
| (Tien et al., 2022) | Occupant activity _ sitting | 0.925 | 0.891 |
|  | Occupant activity _ standing | 0.906 | 0.828 |
|  | Occupant activity _ walking | 0.864 | 0.927 |
|  | Window open | 0.962 | 0.909 |
| (Xiao et al., 2022) | Worker tracking in off-site construction | 0.987 | 0.967 |

### 3.3. Analysis of occupancy detection

Figure 4 presents the actual and detected occupancy curves on the test subset. The model successfully detects the majority of occupancy states, with only a few exceptions. Specifically, out of 968 thermal images in the test subset, only 12 occupied images were not detected. Figure 5 presents the detection results for representative thermal images, corresponding to the scenarios shown in Figure 2. As seen in Figure 5(b)-(f) and Figure 5(j), the model correctly detects occupancy even under challenging conditions such as partial face visibility, side-face views, and faces obscured by a hand. In contrast, Figure 5(g)-(i) show three occupied scenarios that the model failed to detect. In these cases, although the room was indeed occupied, the model incorrectly identified it as unoccupied. This misclassification is likely due to the limited visibility of the occupant face, as specific head orientations (e.g., side-face views) and occupant gestures (e.g., drinking) do not provide sufficient visual information for the model to capture, leading to incorrect identification. These observations suggest that head orientations, occupant gestures, and face visibility play a significant role in influencing the model detection performance.

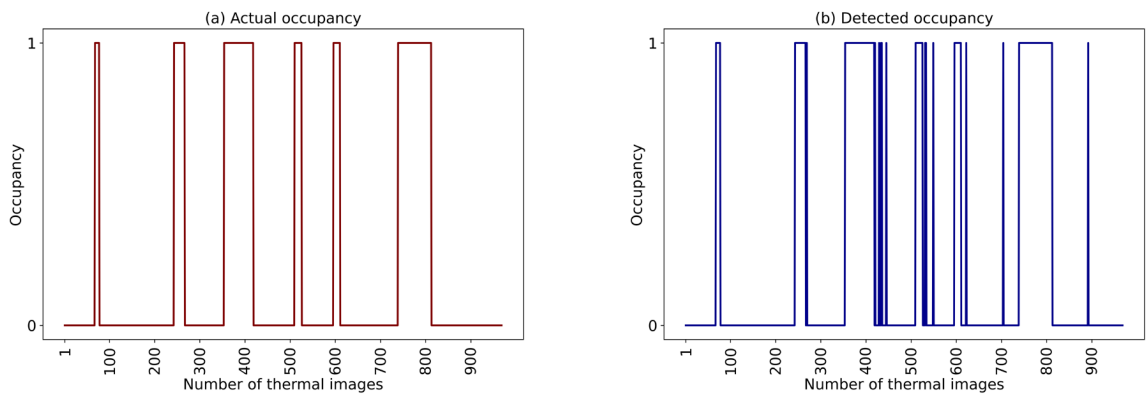

**Figure 4.** Actual and detected occupancy curves on the test subset.

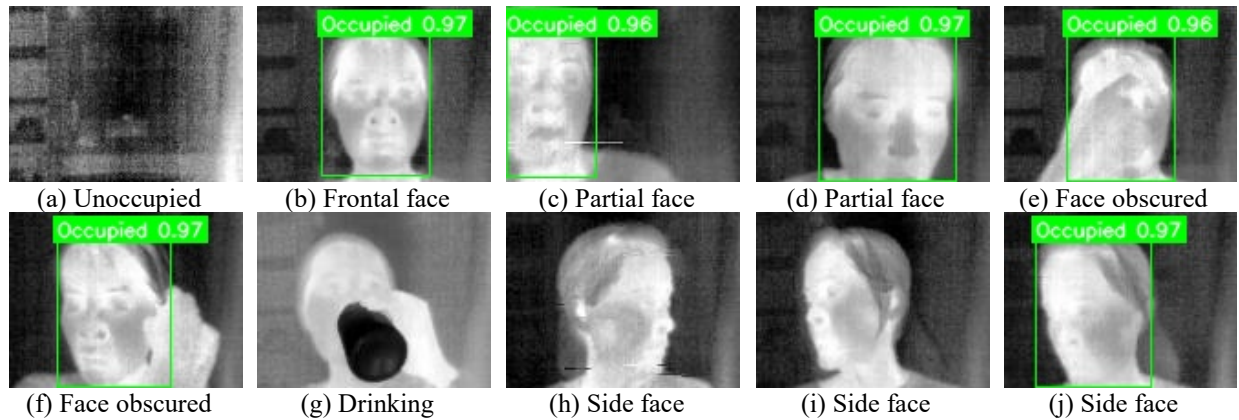

**Figure 5.** Detection results of representative thermal images corresponding to Figure 2.

## 4. CONCLUSION

Occupancy plays an essential role in influencing the energy consumption and operational status of HVAC systems. OCC system has become increasingly recognized in HVAC management since it utilizes actual occupant information to optimize system operation. Reliable occupancy detection is a fundamental requirement for developing an effective OCC system, as it directly determines the on/off status of HVAC equipment. However, most existing CV-based detection models rely on RGB images, which can raise privacy concerns and require high computing resources. To address these challenges, the study developed an occupancy detection model based on low-resolution thermal images combined with CV techniques. A total of 4,836 thermal images were collected for model development, ultimately achieving $precision$, $recall$, $mAP50$, and $mAP50-95$ values close to 1.000 on the test subset. Furthermore, the detection results reveal that head orientations, occupant gestures, and face visibility significantly affect the model performance.

The developed occupancy detection model demonstrates exceptional accuracy, making it highly suitable for integration into OCC systems to determine the on/off status of HVAC systems and potentially achieve energy savings. Additionally, its non-invasive design effectively addresses privacy concerns, enhancing psychological acceptability among building occupants. The use of low-resolution thermal images also reduces the demands on computing resources, thereby improving the practicality and applicability of the model in real-world building environments. The study acknowledges a limitation that most thermal images collected in this study predominantly capture frontal face views, which may not fully represent diverse real-world scenarios. Model accuracy tends to decrease when processing thermal images with varied head orientations, occupant gestures, or limited face visibility. However, these conditions are commonly encountered in practice. To address this, future research should focus on enhancing the model robustness when detecting thermal images with incomplete or minimally visible occupant faces, heads, bodies, and diverse orientations and gestures, thereby further improving its reliability and adaptability across a wide range of real-world applications.

**ACKNOWLEDGEMENTS**
This work was supported by the PhD Student Stipend and Seed Funding for Competitive Projects from the Department of Building and Real Estate at The Hong Kong Polytechnic University.